\documentclass[1p,review]{elsarticle}
\usepackage[tbtags,fixamsmath]{mathtools}
\usepackage{booktabs,multirow,multicol,ctable}

\usepackage{times}
\usepackage{epsfig}
\usepackage{graphicx}
\usepackage{amsmath}
\usepackage{amssymb}
\usepackage[pagebackref=true,breaklinks=true,colorlinks,bookmarks=false]{hyperref}
\usepackage{amsmath,amsfonts,bm}

\def\eqref#1{equation~\ref{#1}}

\def\1{\bm{1}}

\DeclareMathAlphabet{\mathsfit}{\encodingdefault}{\sfdefault}{m}{sl}
\SetMathAlphabet{\mathsfit}{bold}{\encodingdefault}{\sfdefault}{bx}{n}

\newcommand{\E}{\mathbb{E}}

\usepackage{xcolor}
\usepackage{soul}

\IfFileExists{darkmode.tex}{
    \usepackage{pagecolor}
    \definecolor{myfg}{gray}{0.94} 
    \definecolor{mybg}{gray}{0}
    \input{darkmode.tex} 
    \pagecolor{mybg}
    \color{myfg}
}{} 

\def\mypar#1{\vspace{0.15cm}\noindent{\bf #1.}}
\def\myparnovspace#1{\noindent{\bf #1.}}

\def\etal{et al.\ }

\usepackage{amsmath,amssymb,amsbsy,xspace}
\makeatletter
\DeclareRobustCommand\onedot{\futurelet\@let@token\@onedot}
\def\@onedot{\ifx\@let@token.\else.\null\fi\xspace}

\def\eg{{e.g}\onedot, } 
\def\ie{{i.e}\onedot, }

\def\etal{{et al}\onedot}

\def\be{\begin{equation}}
\def\ee{\end{equation}}
\def\bea{\begin{eqnarray}}
\def\eea{\end{eqnarray}}
\def\fig#1{Figure~\ref{fig:#1}}
\def\tab#1{Table~\ref{tab:#1}}
\def\sect#1{Section~\ref{sec:#1}}

\makeatother

\usepackage{url}
\usepackage{subcaption}
\usepackage{multirow}

\usepackage[T1]{fontenc}    
\usepackage{hyperref}      
\usepackage{url}            
\usepackage{booktabs}       
\usepackage{amsfonts}       
\usepackage{xcolor}        

\newcommand\blfootnote[1]{%
  \begingroup
  \renewcommand\thefootnote{}\footnote{#1}%
  \addtocounter{footnote}{-1}%
  \endgroup
}

\journal{Neurocomputing}

\begin{document}

\begin{frontmatter}

\title{Semantics-driven Attentive Few-shot Learning over Clean and Noisy Samples}

\author[odtuaddress]{Orhun Bugra~Baran}
\ead{bugra@ceng.metu.edu.tr}

\author[odtuaddress]{Ramazan Gokberk Cinbis}
\ead{gcinbis@ceng.metu.edu.tr}
\blfootnote{\copyright 2022. This manuscript version is made available under the CC-BY-NC-ND 4.0 license https://creativecommons.org/licenses/by-nc-nd/4.0/}
\address[odtuaddress]{Department of Computer Engineering, Middle East Technical University, 06800 Ankara, Turkey}

\begin{abstract}
Over the last couple of years, few-shot learning (FSL) has attracted significant attention towards minimizing the dependency on labeled training examples. An inherent difficulty in FSL is handling ambiguities resulting from having too few training samples per class. To tackle this fundamental challenge in FSL, we aim to train meta-learner models that can leverage prior semantic knowledge about novel classes to guide the classifier synthesis process. In particular, we propose semantically-conditioned feature attention and sample attention mechanisms that estimate the importance of representation dimensions and training instances. We also study the problem of sample noise in FSL, towards utilizing meta-learners in more realistic and imperfect settings. Our experimental results demonstrate the effectiveness of the proposed semantic FSL model with and without sample noise. The source code can be found at \href{https://github.com/bugrabaran/Semantics-driven-FSL}{https://github.com/bugrabaran/Semantics-driven-FSL}.

\end{abstract}

\end{frontmatter}


\section{Introduction}
\label{sec:intro}

Contemporary supervised learning approaches combined with large training datasets yield excellent
results on various recognition problems. However, a major challenge is learning to model
concepts with limited samples.  Few-shot learning (FSL)~\cite{bart2005,fink2004,feifei2006,lake2011} techniques aim to tackle this
problem by learning to synthesize
effective models based on a few examples.

A major source of motivation for studying FSL is the observation that humans, starting at young
ages, can learn new concepts with limited examples~\cite{landau1988}, ~\cite{merriman1991}, ~\cite{smith2017}. In addition, in most real-world classification problems, such as
object recognition~\cite{liu19largescale}, class distributions can be heavily
long-tailed~\cite{reed2001pareto}. FSL research can be seen as the focused
study of learning to recognize in the low-data regime, which can play a central role in building
semantically comprehensive and rich models.

A variety of FSL approaches have been introduced in recent
years.  Most of the recent work can be summarized as follows:
metric learning-based FSL~\cite{snell2017,sung2017,vinyals2016,koch2015}, generative model based statistical data augmentation~\cite{hariharan2017,wang2018,gao2018,schwartz2018,schonfeld2019,antoniou2017,ridgeway2018}, non-generative
data augmentation~\cite{chen2019,devries2017}, feed-forward classifier synthesis~\cite{gidaris2018,qi2018,qiao2017}, model
initialization for few-shot adaptation~\cite{finn2017,nichol2018,rusu2018}, learning-to-optimize
for FSL~\cite{ravi2016} and memory-based FSL approaches~\cite{santoro2016,sukhbaatar2015}.  In addition,
recent work has highlighted the importance of implementations details in improving and
evaluating FSL
models, including batch normalization details~\cite{bronskill2020}, feature extraction
backbones~\cite{chen2019closer} and pretraining strategies~\cite{chen2020new}. Variations of FSL, such as cross-domain~\cite{xing2019,schwartz2019,fu2019,mu2019,li2020} and
variable-shot~\cite{triantafillou2019} learning have also been introduced.

\begin{figure*}
    \centering
    \subcaptionbox{Spurious background.\label{fig:spurious:bg}}
    {\includegraphics[width=0.33\textwidth]{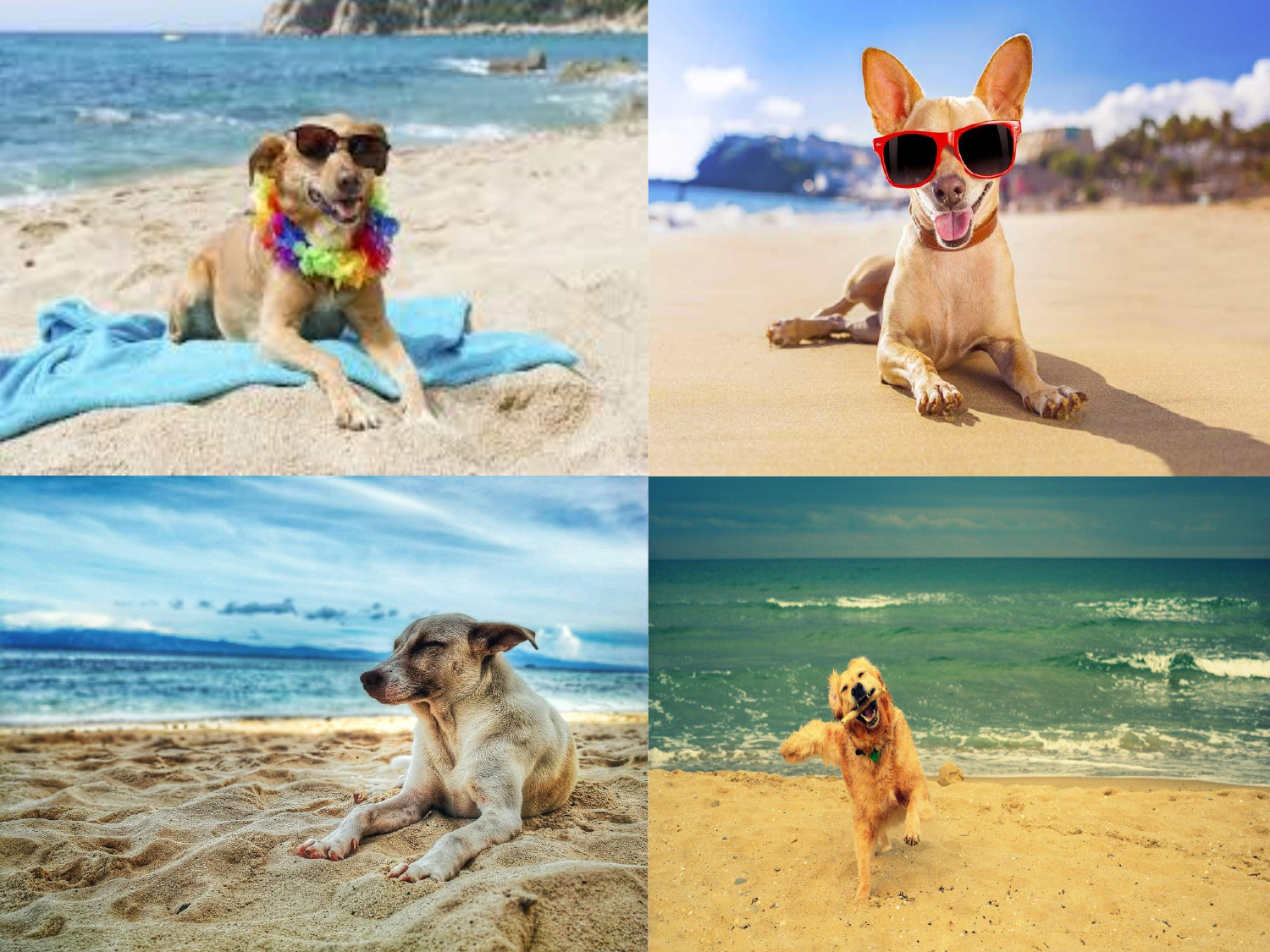}}
    \subcaptionbox{Misleading relations.\label{fig:spurious:fgbg}}
    {\includegraphics[width=0.33\textwidth]{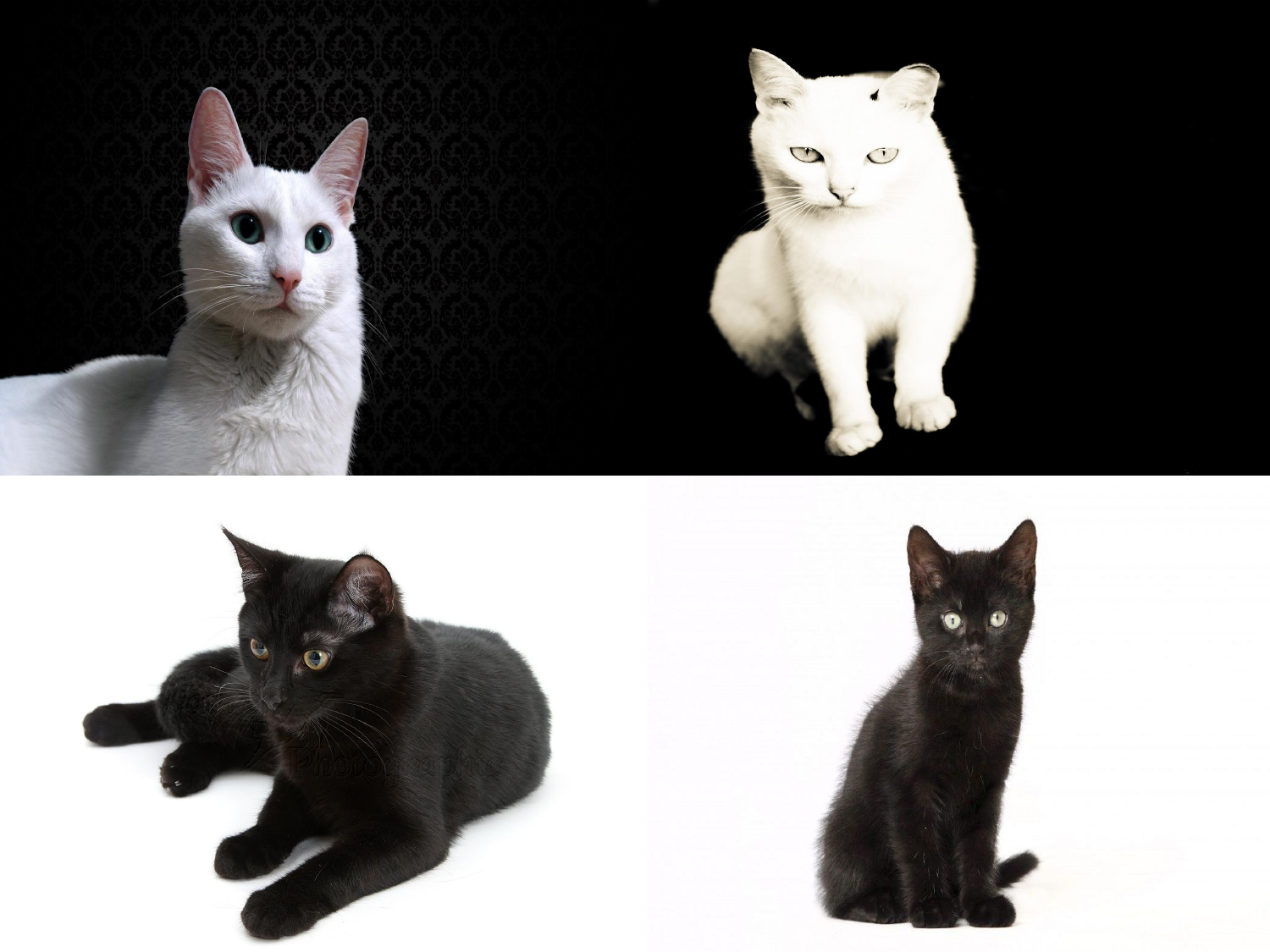}}
    \subcaptionbox{Feature ambiguity.\label{fig:spurious:fg}}
    {\includegraphics[width=0.33\textwidth]{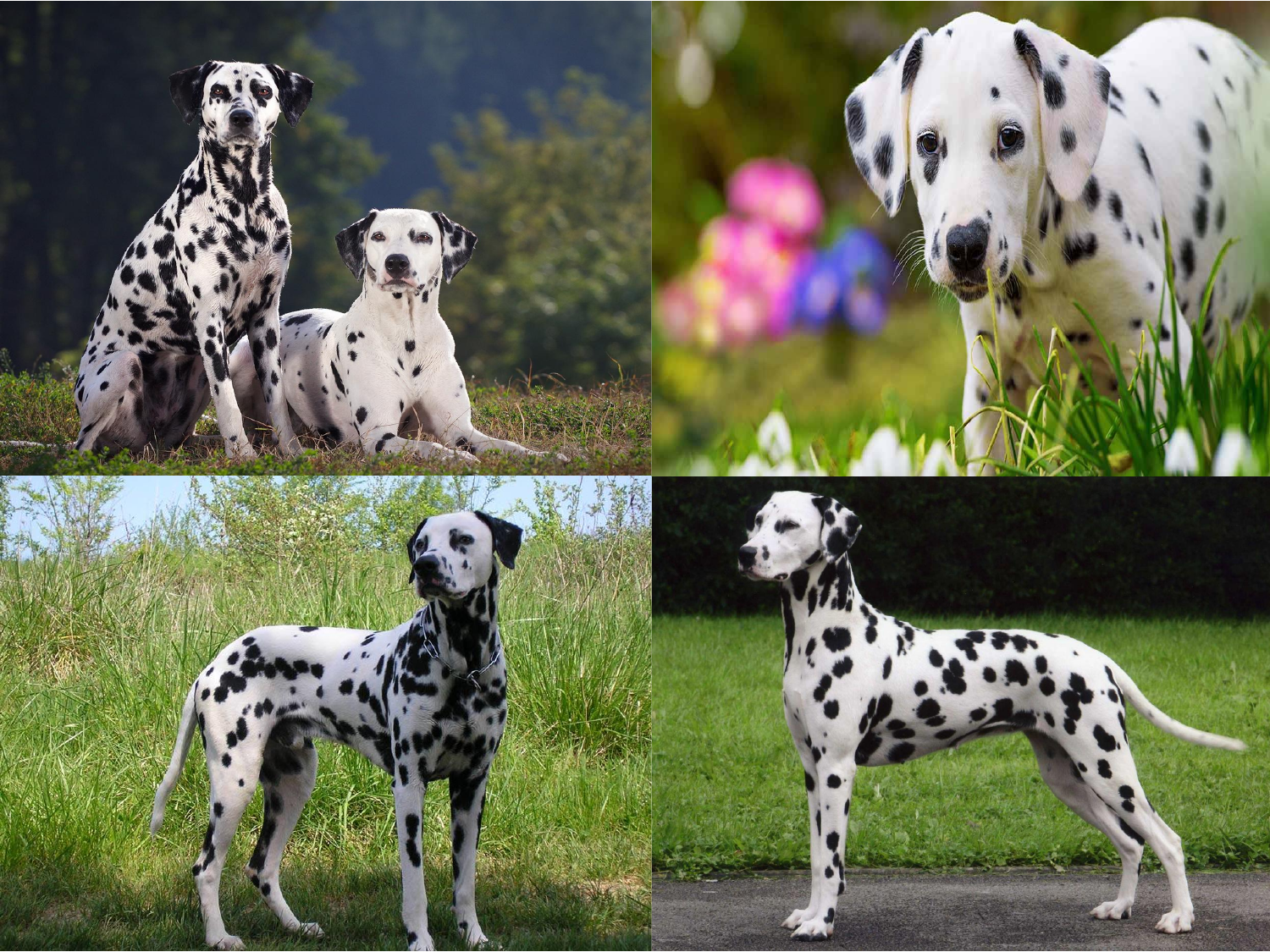}}
    \subcaptionbox{Prototypicality/noise.\label{fig:spurious:representativeness}}
    {\includegraphics[width=0.33\textwidth]{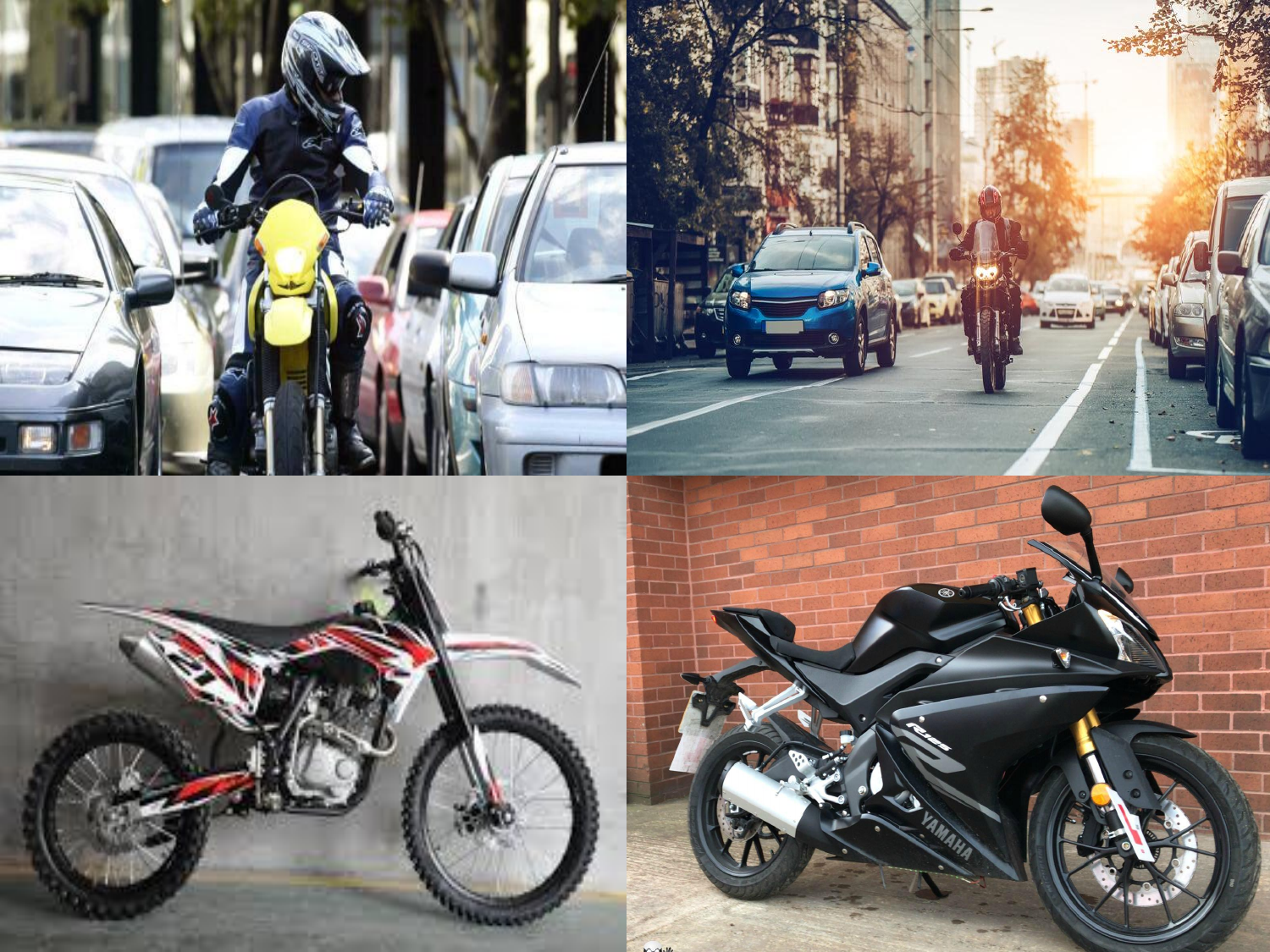}}
    \caption{
    (a) All dog instances appear in a misleadingly consistent beach background context.
    (b) Spuriously consistent background-foreground relation may cause FSL models to ignore other salient features of classes. 
    (c) It is difficult to understand when a consistent foreground texture is informative or misleading, \eg the Dalmatian texture is distinctive but the {\em Dalmatian dog} class but otherwise misleading for the generic {\em dog} class. 
    (d) Some examples can be more {\em prototypical} than the others. 
    A closely related problem is having {\em sample noise} in the training set. \label{fig:spurious}}
\end{figure*}

An inherent difficulty in FSL, independent of the method being used, is the ambiguity resulting from
having few training samples per class. Particularly, it is difficult to figure out whether a cue that
appears consistently in the limited set of examples is truly indicative. 
For example, few-shot samples may
contain misleadingly similar contextual information (\fig{spurious:bg}), 
spurious foreground-background relationships (\fig{spurious:fgbg}), 
or suspiciously consistent foreground features (\fig{spurious:fg}).
Similarly, some samples might actually
be less {\em prototypical}~\cite{rosch1973111} than the others or completely {\em noisy}, due to a number of factors, such as background clutter, viewpoint, occlusions, overall representativeness and sample noise (See \fig{spurious} for further discussion). 
Hence, the FSL model may {\em overfit} 
to incorrect features, misleading or noisy samples, or under-utilize distinctive cues, due to the fundamental difficulty of 
disambiguating spurious cues from the informative ones
purely based on a few examples.

To tackle these fundamental challenges in FSL, we aim to guide the meta-learner via semantic priors,
which we call {\em semantic few-shot learning}.
To this end, we build meta-learning models that can benefit from text-based 
semantic representations of classes of interest when synthesizing target classifiers. For this
purpose, we focus on one of the most popular metric based few-shot learners
Prototypical Networks~\cite{snell2017} (PNs). In the context of PN formulation, 
we introduce semantically-conditioned {\em feature attention} and
{\em sample attention} mechanisms, towards reducing the risk of overfitting to misleading features or samples and 
improving the data efficiency in few-shot learning.

The use of semantic vector-space representations of classes, \ie {\em class embeddings}, is prominent in
{\em zero-shot learning} (ZSL). Most mainstream ZSL approaches learn 
to estimate the
degree of relation between a given input (image) and a class embedding, so that 
previously unseen classes can be recognized purely based on class embeddings, \eg \cite{frome2013,norouzi2013,zhang2016,chen2017,schonfeld2019,xian2019,mishra2018generative,zhu2017}.
In ZSL, class embeddings can be interpreted as class summaries from which
valuable discriminative or generative knowledge can be extracted. In our work, instead of relying
purely on class embeddings for building classification models, we aim to benefit from them
in improving sample efficiency in FSL.

The idea of jointly benefiting from class semantics in FSL has received attention only 
recently. In \cite{xing2019}, semantics-based class prototype priors are
estimated and adaptively combined with the data-driven class prototypes.  \cite{schwartz2019}
extends this model to multiple semantic information sources.
\cite{fu2019} aims to obtain augmented feature representations
based on embedding images into the semantic space and then decoding the sampled semantic
features to get new samples for training purposes. 
\cite{li2020} proposes to reformulate the PN loss based on
task-dependent similarities measured from semantic priors to regulate the margin between classes in
a task.  

The prior work most related to ours is the AM3 model~\cite{xing2019}, which we use as
our starting point.  While AM3 makes a significant step forward by defining class
semantics-based models priors, the approach does not leverage prior semantic knowledge in knowledge
accumulation. 
To this end, we explore the uses of semantic class knowledge more deeply in
the following two main ways. First, we aim to estimate the importance of provided few-shot
samples for each class by evaluating the consistency across samples and semantics. Second, we
estimate per-dimension feature importance factors for the data-driven class prototypes based on
prior knowledge. Third, we define a {\em noisy FSL} problem, where 
some of the support samples can be incorrectly labeled, and evaluate the proposed approach in this setting.
We believe that 
noise-tolerant FSL models can have applications in variety of real-world problems,
\eg the construction of large-vocabulary models over automatically retrieved web samples based on
meta-data where top ranking items tend to be much more coherent yet not without noise;
large-scale fine-grained FSL problems where human annotators are prone to making mistakes;
and on-the-fly construction of models from few interactively provided samples in robotic systems.

Our contributions can, therefore, be summarized as follows: (i) we propose a semantics-driven feature
and sample attention mechanisms to improve FSL data efficiency in a principled way; (ii) we define 
an experimental setting for noisy FSL and 
investigate the applicability of the proposed sample attention approach to this problem; (iii) we
present a detailed empirical analysis toward understanding the effects and dynamics of the
proposed attention mechanisms.  Our quantitative and qualitative experimental results demonstrate
the effectiveness of the proposed semantic FSL model both in clean and noisy settings.

\section{Related work}
\label{sec:relwork}
 
In this section, we first present an overview of mainstream few-shot learning approaches. We then
overview works on the recently emerging topic of semantic FSL. Finally, we briefly
discuss zero-shot learning and its relation to our work.

\mypar{Few-shot learning} The most related mainstream FSL approaches can be summarized within
{\em initialization based}, {\em metric learning based} and {\em generative model based} groups.
Initialization-based FSL methods aim to learn the {\em ideal} initial
model such that the model can perform well even when fine-tuned using just a few examples.
MAML~\cite{finn2017} is arguably the most well-known example of this category. In MAML,
the main idea is to learn the initial model that minimizes the loss of validation samples 
when the initial model is fine-tuned using one or few gradient-based updates.
Several other related and follow-up works exist, such as \cite{nichol2018,rusu2018,li2017,ravi2016}.

In metric learning-based FSL, the goal is to learn a metric space
where the similarity of feature representations of sample pairs can be used to classify pairs
as same class/different class pairs.
One of the most well-known examples of this category is Prototypical Networks~\cite{snell2017}, which we also 
use as the basis of our approach (see \sect{method}). Due to its simplicity and high FSL performance,
many other metric learning FSL approaches have also been introduced, \eg 
\cite{sung2017,vinyals2016,gidaris2018,qi2018}. Despite these explorations, recent works show that a carefully tuned
Prototypical Network can yield state-of-the-art few-shot learning results~\cite{chen2020new,tian_rethinking_2020,feat}. We 
adapt the {\em modernized} PN implementation of Ye \etal~\cite{feat} as the non-semantic FSL baseline in the construction of our models. 

Generative modeling based FSL approaches aim to learn a sample-synthesizing model, which can be used
for augmenting a few-shot training set. For example, \cite{hariharan2017} learns a mapping that can
be used to transform existing train samples into new ones, \cite{zhang2018,gao2018} propose
GAN~\cite{goodfellow2014} based generative models towards synthesizing novel examples.

Another important research direction is learning generalizable feature representations~\cite{tadam,
tian_rethinking_2020, bronskill2020}. The representation generalizability is a major problem in FSL
as backbone networks are utilized on novel classes at test time. \cite{sarn} uses self-attention
over spatial locations to improve representations, similar to non-local networks~\cite{wang2018non},
for a relation-network-based FSL approach. \cite{Yan2019ADA} proposes a transductive FSL approach
that aims to obtain query-specific feature representations via attention mechanisms. These
approaches are orthogonal to ours, as we focus on leveraging semantic priors in a query-independent
way.

Sample noise in FSL is largely an overlooked problem. To the best of our knowledge, the only directly
relevant work is the few-shot text classification approach  of \cite{gao_hybrid_2019}, which looks
into noisy annotations for few-shot relation classification. \cite{gao_hybrid_2019} defines
query-to-support sentence and support samples driven feature-level attention mechanisms.  Our work
and focus fundamentally differs  as we (i) leverage prior knowledge for building 
conditional attention mechanisms, (ii) estimate sample importance in a query-agnostic way, and (iii)
define an experimental protocol for studying the sample noise problem on the mainstream FSL
image classification benchmarks MiniImageNet~\cite{vinyals2016} and TieredImageNet~\cite{tieredimagenet}.

\textbf{Semantic few-shot learning}. Semantic FSL refers to an FSL problem variant where a supplementary
class-wise knowledge source is available. Since such additional knowledge often comes from a
new data modality, semantic FSL is also sometimes referred to as {\em multi-modal FSL}.  There exist
only a few recent works on semantic FSL. 
In a pioneering work, \cite{xing2019} proposes to define class prototypes as convex combinations of
average visual features and transformed semantic priors. \cite{schwartz2019} extends the approach of
\cite{xing2019}, mainly by introducing multiple semantic priors (label, description, or attribute)
jointly to obtain richer semantic information.  In \cite{fu2019}, visual features are mapped into a
semantic space via an encoder, where semantic features belonging to the same class can be
sampled to augment visual features via a decoder model. The approach then performs classification by
using both real and augmented features. Similarly, \cite{schonfeld2019} aims to learn aligned
auto-encoders with reconstruction losses across modalities.  In more recent work, \cite{li2020} uses
semantic prior information with a similarity function to obtain margin scores that are used as an
additional term in a distance-based loss, such as \cite{snell2017}. \cite{peng2019} uses semantic
priors in addition to visual features to obtain two different classification weights for a class. This
work obtains visual feature-based classification weights by l2-normalizing support image features,
averaging them per class, and learning a semantic embedding vector that can be used as a classifier
weight maximizing the inner product between the support image based classification weights.
\cite{ji_reweighting_2021} proposes a model that re-weights support samples according to a generic
weighting function and an auto-encoder that can use either class embeddings or Gaussian noise
vectors for encoding regularization. Among these models, the most closely related one is
AM3~\cite{xing2019}, as we build our models on top of it.  
While nearly all others can be considered
as complementary to ours, instead of being alternatives, we provide empirical comparisons to 
semantic FSL works in Section~\ref{sec:exp}.

\textbf{Zero Shot Learning}. Zero-shot learning aims to build recognition models that can handle
classes with no training examples purely based on prior class semantic knowledge.  Mainstream ZSL
approaches include learning mappings between the space of visual features and the semantic class
representations \cite{frome2013,norouzi2013,zhang2016}, and semantics conditional generative models
\cite{chen2017,schonfeld2019,xian2019,mishra2018generative,zhu2017}.  In our work, we use class
semantics to build attention mechanisms in the presence of a few training examples instead of aiming
to remove the need for training samples completely, as in zero-shot learning.

\section{Method}
\label{sec:method}

In this section, we first provide a formal definition of semantics-driven few-shot learning and
summarize the {\em episodic training} framework, which we embrace in our approach. We then give a
summary of the Prototypical Networks (PN) model~\cite{snell2017} for few-shot learning and 
{\em adaptive cross-modal few-shot learning} (AM3)~\cite{xing2019} model that extends PNs by 
utilizing semantic knowledge to construct model priors. Finally, we present the proposed feature-attention and
sample-attention mechanisms and explain how we integrate them into the PN and AM3 models.

\subsection{Problem definition and training framework}

\def\D{\mathcal{D}}
\def\Ds{\mathcal{D}^\text{s}} 
\def\Dsc{\mathcal{D}^\text{s,c}} 
\def\Dq{\mathcal{D}^\text{q}} 
\def\Dqc{\mathcal{D}^\text{q,c}}
\def\C{\mathcal{C}}

In our work, we focus on few-shot learning of image classification models. The goal is to
estimate a new classification model for a set of target classes ($\C=\{c_1,...,c_n\}$), based on a
limited set of labeled training examples.  In our discussion, an $n$-way classifier is
expressed in terms of a scoring function $f(x;\theta)$ that maps the input $x \in \mathcal{X}$ to an
$n$-dimensional vector, according to the model parameters $\theta$.  In a standard supervised
training problem, a typical way to estimate $\theta$ is to find the model parameters minimizing some
regularized empirical loss function on the train set.

However, in the case of few-shot learning, the main challenge is to estimate a successful
classification model based on a few training samples per class.  It is
fundamentally difficult to achieve generalization based on a few examples in most practical cases. Therefore, a model learned
by minimizing a generic empirical loss function is unlikely to perform well on novel (test) data. To
tackle this problem, the main interest in meta-learning-based FSL is to learn a
meta-learner $\xi(\D;\beta)$ that can take a new limited set $\D$ of training examples and synthesize the
corresponding classification model parameters.  $\beta$ represents the trainable parameters of the meta-learner model $\xi$.

\mypar{Episodic training} A popular approach for training meta-learning models is
{\em episodic training}. The main idea is to construct a set or series of few-shot learning
tasks and update the meta-learning model based on the regularized empirical loss of meta-learned 
classification models. More specifically, at each iteration, a new task $T$ is created by
sampling a subset $\C_T$ of training classes and then sampling training $\Ds_T$ and validation $\Dq_T$ samples
from the whole training set. Each $\Ds_T$ consists of few-shot training samples of classes $\C_T$ and is commonly referred to as 
the {\em support set}. Similarly, each $\Dq_T$ consists of task-specific validation samples and is commonly referred to as
the {\em query set}. The meta-learning, then, is achieved by minimizing the expected empirical loss of
meta-learned models over the pairs of support and query sets:
\begin{equation}
    \min_\beta \E_{T} \left[ \E_{(x,y)\sim \Dq_T} \left[ l(f(x;\theta=\xi(\Ds_T;\beta)),y) \right] \right] 
\end{equation}
where $l(\cdot,y)$ is a classification loss function for label $y$. $\theta$ represents the task-specific model parameters inferred by the meta-learner $\xi$. 
No regularization term is shown for brevity.

\mypar{Semantic FSL} Arguably, the central premise of meta-learning is to learn domain-specific
inductive biases better than what can be provided by general-purpose supervised learning formulations.
However, small-sized training sets can be inherently misleading and/or
ambiguous due to spurious patterns, as previously illustrated in \fig{spurious}.  In this respect,
prior knowledge about classes can be a crucial source of information for overcoming the
limitations of few-shot training sets. In our work, we presume that semantic knowledge about each
class $c$ is a $d_e$-dimensional vector, represented by $\psi_c$.  Therefore, the meta-learner $\xi$ can also access
semantic class embeddings during both training and testing. In the following parts, we define the baseline and proposed meta-learner $\xi$ variants 
in formulations for inferring task-specific parameters $\theta$ from a set of support samples and semantic priors.

\def\protoPN{\theta^\text{PN}}
\def\protoAM3{\theta^\text{Prior}}
\def\protoInit{\theta^\text{0}}
\def\protoFeatAtt{\theta^\text{FeatAtt}}
\def\protoSampleAtt{\theta^\text{SampleAtt}}
\def\protoCombined{\theta^\text{Combined}}
\def\attAlpha{\eta_\alpha} 
\def\attFeature{\eta_\text{FeatAtt}} 
\def\attSample{\eta_\text{SampleAtt}} 
\def\attSampleUnnormvis{\gamma_{vis}} 
\def\attSampleUnnormsem{\gamma_{sem}} 
\def\transPrior{\tau_\text{Prior}}

\begin{figure*}
    \centering
    \includegraphics[width=0.80\linewidth]{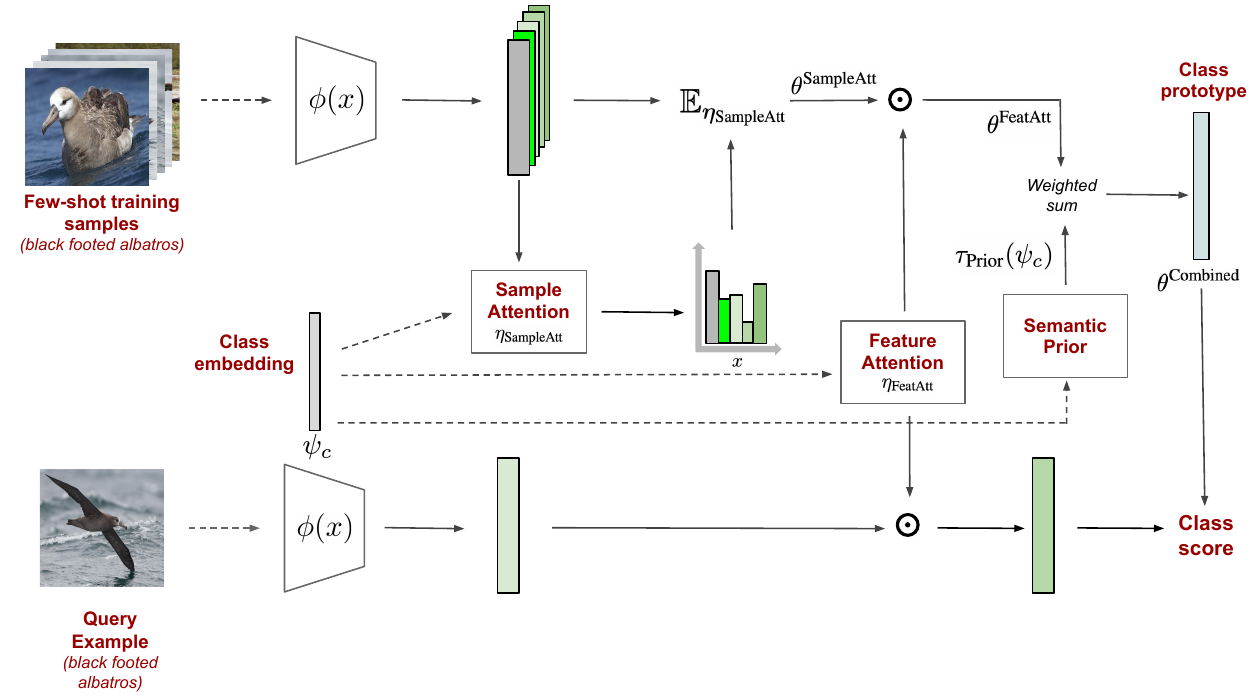}
    \caption{Summary of the proposed meta-learner ($\xi$), illustrating the process of a new class prototype estimation 
    in the 5-shot setting. Arrows indicate the input dependencies across model components.  Dashed lines are used to indicate externally provided data input.
    The sample attention model ($\protoSampleAtt$) uses class embeddings ($\psi_c$) to assign relative importance weights to the provided support examples.
    The feature attention model ($\protoFeatAtt$) similarly uses class embeddings to estimate task-specific feature dimension scaling coefficients.
    The final model ($\protoCombined$) is obtained by combining the sample and feature attention-driven task parameters with the semantics-only prior ($\protoAM3$). The resulting model is applied to the test input(s). ~\label{fig:method}}  
\end{figure*}

\subsection{Semantics-driven attention mechanisms}

We build our semantics-driven attention mechanisms on top of the semantic few-shot
learning model AM3, which is based on the prototypical network (PN) model. Below we first summarize the PN and AM3
models and then present our approach in its context.

\mypar{Prototypical networks and AM3} The core idea in PN is to estimate {\em class prototypes} based on
train samples provided in a task and then perform classification based on query-to-prototype similarities.
In PNs, a class prototype $\protoPN$ for a class $c$ is obtained by averaging the support sample representations:
\begin{equation}
    \protoPN(c) = \E_{\Dsc} \left[ \phi(x) \right]
\end{equation}
where $\phi(x)$ is the feature embedding function parameterized by (a subset of) $\beta$ that is being trained as part of the PN model.
$\phi(x)$ is ResNet12 in our experiments (see \sect{exp}).
$\Dsc$ refers to the subset of examples belonging to class $c$ within a given support set.
In the de facto standard PN formulation, the classification score $f$ for 
class $c$ is given by the negative Euclidean distance between the feature-space embedding of the input
and the corresponding class prototype:
\begin{equation}
    \left[f(x)\right]_c = -\| \phi(x) - \protoPN(c) \|^2 .
\end{equation}
The parameters $\beta$ are estimated by minimizing cross-entropy loss of query samples over the episodes.

The AM3 model aims to improve the PN model by using semantic knowledge about classes as prototype priors. 
More specifically, AM3 redefines a class prototype $\protoAM3$ as the weighted combination of transformed class semantic embeddings
and feature averages:
\begin{equation}
    \protoAM3(c) = \alpha \protoPN(c)  + (1-\alpha) \transPrior(\psi_c)
\end{equation}
where $\transPrior$ is a trainable transformation that maps the semantic class embeddings to the space of class prototypes.
$\transPrior$ is parameterized by $\beta$.

The AM3 model, therefore, can be interpreted as a way to build classification model priors in the PN
formulation. Consistent with the experimental results, such a prior is particularly valuable
in the case of one-shot learning, where the training data size is at its extreme minimum. 
We propose and explore two novel ways to leverage semantic information in a more expressive way toward 
tackling the inherent difficulties in few-shot learning: 
(i) {\em sample attention}, (ii) {\em feature attention}.
Below we provide the details of the proposed mechanisms, which act as the components of the whole model, a scheme of which
can be found in \fig{method}.

\mypar{Sample attention} We observe that, in the original PN model, each class prototype is defined as a plain
average of support examples. However, the information content of samples can vary
greatly due to a number of factors, including background clutter, viewpoint and occlusion. Towards 
estimating the prototypicality of training samples, we introduce a sample attention mechanism into the
final model. Ultimately, we aim to build a model that can estimate the importance of each sample 
based on the compatibility of samples and prior semantic information. Therefore, we define sample attention
module $\attSample(x,c)$ as a function that computes the normalized attention scores of a sample $x \in \Dsc$
in the context of support sample set $\Dsc$, conditioned on the class semantic embedding vector $\psi_c$:
\begin{equation}
    \attSample(x,c) = \frac{\exp(\attSampleUnnormvis(\phi(x))^\top \attSampleUnnormsem(\psi_c))}{\sum_{x^\prime \in \Dsc} \exp(\attSampleUnnormvis(\phi(x^\prime))^\top \attSampleUnnormsem(\psi_c))}
\end{equation}
where $\attSampleUnnormvis$ and $\attSampleUnnormsem$ are trainable models that are used to obtain visual and semantic feature embeddings.
In our experiments, $\attSampleUnnormvis$ and $\attSampleUnnormsem$ are implemented as MLPs that take $\phi(x)$ and $\psi_c$ respectively and return embeddings 
whose inner products yield the attention scores. We note that $\attSample$ defines a distribution 
over the support samples, which we use to re-define a class prototype:
\begin{equation}
    \protoSampleAtt(c) = \E_{\attSample(x,c)} \left[ \phi(x) \right] ,
\end{equation}
which amounts to computing the attention-weighted average of sample features. This sample attention mechanism also naturally 
handles potential sample noise in FSL, which we explore in Section~\ref{sec:exp}.

\mypar{Feature attention} The motivation in feature attention is to tackle the problems resulting from having too
few examples. Overall, the goal is to enhance or attenuate certain prototype dimensions as a function of semantic 
class embeddings. For this purpose, we introduce the feature attention function $\attFeature$ and re-define the 
class prototype as follows:
\begin{equation}
    \protoFeatAtt(c) = \attFeature(\psi_c) \odot \protoInit(c)
\end{equation}
where $\protoInit(c)$ refers to averaging based class prototypes, which can either be the vanilla PN prototype ($\protoPN$)
or the sample attention based prototype ($\protoSampleAtt$). $\odot$ represents the Hadamard product operator.
Here, the trainable feature attention function $\attFeature$ predicts per-dimension scaling coefficients as a 
function of class embeddings. The feature attention output is applied to both the prototype and the query, resulting in the following
scoring function:
\begin{equation}
    \left[f(x)\right]_c^\text{FeatAtt} = -\| \attFeature(\psi_c) \odot \phi(x) - \protoFeatAtt(c) \|^2 .
\end{equation}

\mypar{The final model} We build our final model by integrating  our semantic-driven sample and feature attention mechanisms
into the semantic FSL framework defined by the AM3 model. More specifically, in the final {\em combined attention} model,
we first estimate sample attention based class prototypes $\protoSampleAtt$ and then update them into $\protoFeatAtt$
using the feature attention model. We obtain the final class prototypes, which we call $\protoCombined$, by computing the 
$\alpha$-weighted combination of the attention-driven prototypes and the pure semantic embedding based prototypes given by $\transPrior(\psi_c)$:
\begin{equation}
        \protoCombined(c) = \alpha  \attFeature(\psi_c) \odot \protoSampleAtt(c) + (1-\alpha) \transPrior(\psi_c) .
\end{equation}
The combined attention model corresponds to the scheme presented in \fig{method}.

\section{Experiments}
\label{sec:exp}

In this section, we first explain our experimental setup and our implementation details. We then
present our main experimental results and analyses. Finally, we present and discuss our results.

\subsection{Experimental setup}
\label{sec:expsetup}

Prior works~\cite{chen2019closer,bronskill2020,chen2020new} show that
implementation details, such as (batch)
normalization schemes, backbones, hyper-parameter tuning strategies, data augmentation schemes,
can make a great impact on few-shot learning results.
Therefore, 
we
systematically tune the hyper-parameters, including learning rates, number of iterations, and
dropout rates on the validation sets for all results that we report based on our experiments.
Below we provide additional details regarding our experimental setup and our efforts to make fair
comparisons.

\myparnovspace{Datasets} We evaluate our model on the MiniImageNet~\cite{vinyals2016} and
TieredImageNet~\cite{tieredimagenet} datasets. The MiniImageNet dataset is a
subset of ImageNet dataset~\cite{imagenet} with 100 classes and 600 images per class.
We use the split of Ravi and Larochelle~\cite{ravi2016} for MiniImageNet where the 64, 16 and 20 classes are 
used as the train, validation and test subsets, respectively.
TieredImageNet is a separate benchmark based on ImageNet~\cite{imagenet},
with 351, 97, 160 classes for training, validation and testing, respectively. 
In all our experiments, we use Glove~\cite{pennington2014} vectors of class names to extract semantic embeddings.
Following Xing \etal~\cite{xing2019}, we use the Common Crawl version trained on 840B tokens with 300-dimensional embeddings.

\mypar{Evaluation} We report our test set results over 10000 random tasks and report the average accuracy and $95\%$
confidence interval scores.  In all k-shot experiments, we use 15 query examples for each of the n
classes in an episode.  We execute all of our experiments using the same batch structure for
consistency across the experiments.  

\mypar{Backbone architecture} Since our primary goal is to evaluate the potential value of proposed semantics driven 
attention mechanisms over a contemporary baseline, we use the same ResNet-12 architecture as the feature
extraction backbone throughout our experimental results. 
We use Batch Normalization~\cite{ioffe2015} only in backbone layers, using the
{\em eval} mode~\cite{paszke2019} for meta-testing to avoid {\em accidental} transductive setting,
which is known to potentially result in misleadingly better few-shot learning
results~\cite{bronskill2020}. 

\mypar{Backbone pretraining} Following Ye \etal~\cite{feat} we use supervised pretraining for ResNet-12
backbone and the exact details can be found in Ye \etal~\cite{feat}.  After the pretraining, we employ a two-staged training approach. In the first stage, we train everything except the backbone with a learning rate
$0.1$, momentum $0.9$ and weight decay of $0.0005$ for 200 epochs for MiniImageNet and 50 epochs for
TieredImageNet, where an epoch contains 100 episodes. After every epoch, we validate our results by
using 600 episodes for MiniImageNet and TieredImageNet. After we complete the 200 epochs and 50
epochs for MiniImageNet and TieredImageNet, respectively we train the whole model end to end for another
200 epochs. In the second stage, the learning rate for the backbone is selected to be $0.002$ and $0.02$
for the rest of the network. After every 40 epochs we halve the learning rates. We use SGD as the
optimizer for every model and experiment.

\mypar{PN implementation} Our code base is built upon the PN implementation of Ye \etal~\cite{feat}.  For a
a fair comparison, therefore, we report the results that we obtain for PN and FEAT~\cite{feat} using the
publicly available official source codes for FEAT~\cite{feat}.  Similar to \cite{tadam}, we use distance
scaling and found that it is better to divide the distances by 32 for MiniImageNet and 16 for
TieredImageNet based on the validation results. 

\mypar{AM3 implementation} In our re-implementation of AM3~\cite{xing2019}, 
we have obtained the highest validation accuracy scores using
$0.4$ as the dropout rate for the fully-connected (FC) layers
on both MiniImageNet and TieredImageNet datasets. 
We utilize the same Euclidean scaling in AM3 as in PN to
obtain fair results. 
Overall, our implementation significantly improves the performance of the
baseline AM3 model compared to the scores originally reported in Xing \etal~\cite{xing2019}.

\mypar{Proposed model implementation} In the proposed model, we use $2$ FC layers to encode the visual and semantic information for
the sample attention module. These FC layers have the structure of FC-Dropout-ReLU-FC, and the dimensions
are reduced to $32$. The dropout probabilities are selected to be $0.2$ and $0.6$ for visual and
semantic branches, respectively, using the validation sets. For the feature attention module, we use
2-layer FC module with the structure of FC-Softmax-FC and the dimensionality is first reduced to
$32$ and then increased to the feature dimensionality, which is $640$ for ResNet-12. Finally, we
also add the semantic prototype branch of AM3~\cite{xing2019} exactly as it is, with the only
difference of keeping the $\alpha$ a fixed hyper-parameter based on the validation set instead of
predicting it from the semantic prior.  

\mypar{Source code} Training of our final model takes approximately 4 hours on a single V100 GPU. The source code can be found at \href{https://github.com/bugrabaran/Semantics-driven-FSL}{https://github.com/bugrabaran/Semantics-driven-FSL}.

\subsection{Main results}

This section presents our main results, ablative studies, and experimental analysis of the
proposed model on few-shot learning benchmarks. We then present a comparison to state-of-the-art
approaches. Finally, we evaluate the approach on the more challenging 10-way and 15-way few-shot learning setups.

\mypar{Baselines} \tab{tableMain} presents our main results for 1-shot or 5-shot, 
5-way classification, including baselines that are carefully tuned in the same way to make fair
comparisons. First of all, we validate our PN and AM3 baselines. For this purpose, we compare our PN
and AM3 results (lower part) to the results reported in the original AM3 work~\cite{xing2019} (upper
part of \tab{tableMain}).  Overall, our results for both baselines are significantly better than the
originally reported ones.  Noticeably, our PN results on MiniImageNet are higher by nearly $7$ and
$4$ points in 1-shot and 5-shot cases, respectively. Similarly, our AM3 results are higher for
nearly $4$ and $5$ points in 1-shot and 5-shot cases. We observe even larger
improvements for both baselines on TieredImageNet. These results re-highlight the importance of
implementation details in FSL and validate the strength of our main baselines. A major factor in
obtaining strong baselines is {\em backbone pretraining}, for which we present additional results in
the context of our work later in this section.

\mypar{Main results and ablative experiments} The very last row of \tab{tableMain} contains our main
results using the final combined attention models. The preceding lines present the results for the
ablated versions of the model with only sample attention or only feature attention based AM3
extensions. We note that in the case of 1-shot learning, sample attention has no difference by
definition. From the results, first, we observe that sample attention in the case of 5-shot learning slightly degrades
by $0.3$ points on MiniImageNet but improves on TieredImageNet by nearly $0.5$ points. Second, feature
attention improves the AM3 model by approximately $2$ points for 1-shot and slightly degrades
by $0.5$ points for 5-shot on MiniImageNet.  We observe similar patterns on TieredImageNet.

Looking into the final combined attention model results in \tab{tableMain}, we observe consistent
improvements in all cases.  The proposed combined attention improves 1-shot learning from $63.62$
(PN) and $67.55$ (AM3) to $69.76$ on MiniImageNet. Similarly, 5-shot results improve from $78.37$
(PN) and $80.22$ (AM3) to $81.19$. We observe similar improvements on TieredImageNet: 1-shot results
improve from $67.58$ (PN) and $72.60$ (AM3) to $72.69$, and 5-shot results improve from $84.71$ (PN)
and $84.59$ (AM3) to $85.29$.

\begin{table*}[t]
\centering
    \caption{Evaluation of our sample-attention, feature-attention and combined-attention models with comparisons to PN and AM3 as reported in \cite{xing2019} as well as what we obtain by adapting the PN implementation of Ye \etal~\cite{feat}. All results are in the 5-way classification setting, based on ResNet-12 backbone. We use the same model selection policy in all our experiments.}
\resizebox{.8\linewidth}{!}{%
\begin{tabular}{lllll}
\toprule
         & \multicolumn{2}{c}{MiniImageNet}                & \multicolumn{2}{c}{TieredImageNet}              \\ 
 \multicolumn{1}{c}{Model}                                 & \multicolumn{1}{c}{1-Shot} & \multicolumn{1}{c}{5-Shot} & \multicolumn{1}{c}{1-Shot} & \multicolumn{1}{c}{5-Shot}                 \\ \hline
\multicolumn{5}{c}{Results from \cite{xing2019}}                                                                              \\ 
PN \cite{snell2017}      & 56.52 $\pm$ 0.45              & 74.28 $\pm$ 0.20              & 58.47 $\pm$ 0.64              & 78.41 $\pm$ 0.41              \\
AM3 \cite{xing2019} & 65.21 $\pm$ 0.30              & 75.20 $\pm$ 0.27              & 67.23 $\pm$ 0.34              & 78.95 $\pm$ 0.22              \\ \hline

    \multicolumn{5}{c}{Our results}                                                           \\ 
PN \cite{snell2017}         & 63.62 $\pm$ 0.23          & 78.37 $\pm$ 0.21          & 67.58 $\pm$ 0.22          & 84.71 $\pm$ 0.17          \\
AM3 \cite{xing2019}    & 67.55 $\pm$ 0.24          & 80.22 $\pm$ 0.18          & 72.60 $\pm$ 0.21          & 84.59 $\pm$ 0.18          \\
Sample attention (ours)                  & 67.55 $\pm$ 0.24          & 79.93 $\pm$ 0.17          & 72.60 $\pm$ 0.21          & 85.02 $\pm$ 0.16          \\
Feature attention (ours)                 & 69.76 $\pm$ 0.21          & 79.84 $\pm$ 0.15          & 72.69 $\pm$ 0.20          & 84.24 $\pm$ 0.16          \\
Combined (ours)                              & \textbf{69.76 $\pm$ 0.21} & \textbf{81.19 $\pm$ 0.18} & \textbf{72.69 $\pm$ 0.20} & \textbf{85.29 $\pm$ 0.17} \\ 
\bottomrule
\end{tabular}
}
\label{tab:tableMain}
\end{table*}

Noticeably for 5-shot learning on TieredImageNet, the performance gaps between PN, AM3, and our
semantic FSL models are smaller than those on MiniImageNet.  This can be due to the fact that the backbone is
pretrained with a more diverse training set, which is likely to yield better feature representations, 
make few-shot inference tasks less challenging, and reduce the need for leveraging semantic priors.  Consistently, we
also observe that FSL models typically yield higher results on the test set of TieredImageNet compared to MiniImageNet.

\begin{table*}[t]
\centering
\caption{Comparison of PN and our approach with and without pretraining of the ResNet-12 backbone on MiniImageNet.}
\resizebox{.8\linewidth}{!}{%
\begin{tabular}{lllll}
\toprule
\multicolumn{1}{c}{}                  & \multicolumn{4}{c}{MiniImageNet}                                                                  \\ 
    \multicolumn{1}{c}{Model}                                           & \multicolumn{1}{c}{1-Shot Val} & \multicolumn{1}{c}{1-Shot Test} & \multicolumn{1}{c}{5-Shot Val} & \multicolumn{1}{c}{5-Shot Test}            \\ \hline
PN w/o Pretraining \cite{snell2017}  & 43.43 $\pm$ 0.67          & 43.70 $\pm$ 0.24          & 71.90 $\pm$ 0.66          & 69.67 $\pm$ 0.22          \\
Ours w/o Pretraining                       & 64.55 $\pm$ 0.70          & 61.57 $\pm$ 0.23          & 72.46 $\pm$ 0.68          & 68.78 $\pm$ 0.19          \\
\midrule
PN with Pretraining \cite{snell2017} & 68.73 $\pm$ 0.68          & 63.62 $\pm$ 0.20          & 80.88 $\pm$ 0.67          & 78.37 $\pm$ 0.22          \\
Ours with Pretraining                      & \textbf{76.44 $\pm$ 0.65} & \textbf{69.76 $\pm$ 0.21} & \textbf{85.02 $\pm$ 0.65} & \textbf{81.19 $\pm$ 0.18} \\ 
\bottomrule
\end{tabular}
}
\label{tab:tablePretrain}
\end{table*}

\mypar{Importance of backbone pretraining} In Table \ref{tab:tablePretrain} we inspect how
pretraining and backbone quality affects the few-shot learning performance. Here we use both PN and
our approach to see the effect in both uni-modal and multi-modal approaches. Models without
pretraining are trained in an end-to-end fashion. We train the models using the same number of epochs in both cases to create a fair comparison. All of the models are trained for 400
epochs and the results reported based on best validation score. As can be seen from Table
\ref{tab:tablePretrain}, model pretraining improves both approaches very significantly and is
crucial for achieving state-of-the-art performance. An interesting result is the comparison of PN
and our approach for the $1$-shot setting without pretraining since it highlights the significance of
semantic information when the backbone quality and/or visual feature quality is low. As the backbone
becomes better with pretraining, the gap becomes smaller. 

\begin{table*}[t]
\centering
\caption{Few-shot classification accuracy on the test set of MiniImageNet for uni-modal (non-semantic) and multi-modal FSL approaches using ResNet-12 backbones. * indicates the experimental results we obtain based on the implementation from Ye \etal~\cite{feat}.}
\begin{tabular}{llll}
\toprule

     & \multicolumn{2}{c}{MiniImageNet}                    \\
\multicolumn{1}{c}{Model}    &   \multicolumn{1}{c}{1-Shot} & \multicolumn{1}{c}{5-Shot}                     \\ \hline
\multicolumn{3}{c}{Uni-modal few-shot learning baselines}                                                                     \\ \hline
Prototypical Networks*         \cite{snell2017}                         & 63.62 $\pm$ 0.23              & 78.37 $\pm$ 0.21              \\
TADAM                          \cite{tadam}                             & 58.56 $\pm$ 0.39              & 76.65 $\pm$ 0.35              \\
STANet                         \cite{Yan2019ADA}                        & 58.35 $\pm$ 0.57              & 71.07 $\pm$ 0.39              \\
MetaOptNet                     \cite{metaoptnet}                        & 62.64 $\pm$ 0.61              & 78.63 $\pm$ 0.46              \\
FEAT*                          \cite{feat}                              & 65.38 $\pm$ 0.20              & 77.79 $\pm$ 0.15              \\ 
LMPNet                   \cite{lmpnet}                            & 62.74 $\pm$ 0.11              & 80.23 $\pm$ 0.52               \\
FRN                      \cite{frn}                               & 66.45 $\pm$ 0.19              & 82.83 $\pm$ 0.13             \\
IEPT                     \cite{iept}                              & 67.05 $\pm$ 0.44              & \textbf{82.90 $\pm$ 0.30}    \\ \hline
\multicolumn{3}{c}{Multi-modal few-shot learning baselines}                                                                         \\ \hline
RIN                            \cite{ji_reweighting_2021}                               & 56.92 $\pm$ 0.81			   & 75.62 $\pm$ 0.62              \\
ACAM                           \cite{acam}                              & 66.43 $\pm$ 0.57              & 75.74 $\pm$ 0.48              \\
AM3 + TRAML					   \cite{li2020}                            & 67.10 $\pm$ 0.52              & 79.54 $\pm$ 0.60              \\
AM3*                           \cite{xing2019}                          & 67.55 $\pm$ 0.24              & 80.22 $\pm$ 0.18              \\
Ours                                                                    & \textbf{69.76 $\pm$ 0.21}     & 81.19 $\pm$ 0.18              \\ 
\bottomrule
\end{tabular}
\label{tab:tableSotaMini}
\end{table*}

\begin{table*}[t]
\centering
    \caption{Few-shot classification accuracy on the test set of TieredImageNet for uni-modal (non-semantic) and multi-modal FSL approaches using ResNet-12 backbones. * indicates the experimental results we obtain based on the implementation from Ye \etal~\cite{feat}.}
\begin{tabular}{lll}
\toprule
     & \multicolumn{2}{c}{TieredImageNet}                    \\
\multicolumn{1}{c}{Model}    &   \multicolumn{1}{c}{1-Shot} & \multicolumn{1}{c}{5-Shot}                     \\ \hline

\multicolumn{3}{c}{Uni-modal few-shot learning baselines}                                                \\ \hline
Prototypical Networks* \cite{snell2017}           & 67.58 $\pm$ 0.22              & 84.71 $\pm$ 0.19              \\
Relation Net           \cite{sung2018}            & 54.48 $\pm$ 0.93              & 71.32 $\pm$ 0.78              \\
MAML                   \cite{finn2017}            & 51.67 $\pm$ 1.81              & 70.30 $\pm$ 0.08              \\
MetaOptNet             \cite{metaoptnet}          & 65.99 $\pm$ 0.72              & 81.56 $\pm$ 0.63              \\
LEO                    \cite{rusu2018}            & 66.33 $\pm$ 0.05              & 81.44 $\pm$ 0.09              \\
FEAT*                   \cite{feat}               & 70.53 $\pm$ 0.22              & 84.71 $\pm$ 0.15              \\
LMPNet            \cite{lmpnet}             & 70.21 $\pm$ 0.15              & 79.45 $\pm$ 0.17               \\
FRN               \cite{frn}                & 71.16 $\pm$ 0.22              & 86.01 $\pm$ 0.15             \\
IEPT              \cite{iept}               & 72.24 $\pm$ 0.50              & \textbf{86.73 $\pm$ 0.34}     \\ \hline
\multicolumn{3}{c}{Multi-modal few-shot learning baselines}                                                                     \\ \hline
ACAM                   \cite{acam}                & 67.89 $\pm$ 0.69              & 79.23 $\pm$ 0.52              \\
AM3*                   \cite{xing2019}            & 72.60 $\pm$ 0.21              & 84.59 $\pm$ 0.15              \\
Ours                                              & \textbf{72.69 $\pm$ 0.21}     & 85.24 $\pm$ 0.16              \\ 
\bottomrule
\end{tabular}
\label{tab:tableSotaTiered}
\end{table*}

\mypar{Comparison to the state-of-the-art} 
Although comparing classification results across models with different implementation details is
particularly problematic in few-shot learning as previously discussed and shown in the literature, 
it is still of interest to show how well our results are in general compared to the relevant state-of-the-art
methods.  Table \ref{tab:tableSotaMini} and Table \ref{tab:tableSotaTiered} present the results of
ResNet-12 based FSL methods (upper half) and semantic FSL methods (lower half) on MiniImageNet and TieredImageNet
datasets, respectively. While we exclude completely incomparable results, such as those based on different backbones
or transductive models, the results need to be interpreted with caution due to remaining implementation and model selection
differences. From the tables, we first observe that our modernized PN baselines appear to be strong
baselines, outperforming many other more recently proposed approaches. Second, our final model
seems to outperform several methods in FSL and semantic FSL categories on both MiniImageNet and TieredImageNet
datasets. Not surprisingly, the most significant improvements are observed in the 1-shot settings on both datasets compared to 
the uni-modal (non-semantic) approaches since a single training sample is often insufficient, and
semantic prior becomes most valuable in this setting. We also observe that the results reported for the uni-modal models 
in Zhang \etal~\cite{iept} and Wertheimer \etal~\cite{frn} are higher compared to ours in the $5$-shot setting. Here, we note 
that the approach in \cite{iept} has fundamental differences, including a rotation based
augmentation scheme that provides orientation-enriched prototypes and additionally two integrated self-supervised training tasks, which render the results not directly comparable.
Similarly, the approach in \cite{frn} differs fundamentally as it relies on spatially-structured convolutional feature representations instead 
of vector-space embeddings and a feature reconstruction driven class prediction scheme designed for the convolutional features.
Overall, the results highlight the potential value of semantic priming to address the limitations of few-shot learning partially.

\mypar{10-way and 15-way few-shot learning} In Table~\ref{tab:table10way} and Table~\ref{tab:table15way}, we compare
PN, AM3, and the proposed approach for the more challenging 10-way and 15-way settings, respectively. As expected, the
performances of all models decrease as the way count increases, especially in comparison to the commonly studied $5$-way setting. Our model outperforms both PN and
AM3 in the 1-shot setting, in both 10-way and 15-way experiments. In the 5-shot setting, although our approach obtains nearly $1\%$ better
accuracy on validation sets, it falls $0.5$ points behind AM3 on test sets. This points to a less
than the ideal correlation between the validation and test set performances. This is likely to degrade
as the way count increases with a fixed-sized validation (or test) set since the overlap across tasks increases and 
the variance across tasks reduce with larger way settings, reducing the overall richness and reliability of the evaluation.

\begin{table*}[t]
\centering \caption{Evaluation of 10-way few-shot classification on MiniImageNet.}
\resizebox{.7\linewidth}{!}{%
\begin{tabular}{lcccc}
\toprule 
    \multicolumn{1}{c}{} & \multicolumn{4}{c}{MiniImageNet (10-way)}\tabularnewline
\midrule 
Model & 1-Shot Val  & 1-Shot Test  & 5-Shot Val  & 5-Shot Test \tabularnewline
\midrule 
PN \cite{snell2017}  & 53.61 $\pm$ 0.60  & 46.04 $\pm$ 0.13  & 70.30 $\pm$ 0.39  & 66.03 $\pm$ 0.10 \tabularnewline
AM3 \cite{xing2019}  & 61.46 $\pm$ 0.55  & 51.37 $\pm$ 0.12  & 72.92 $\pm$ 0.43  & \multicolumn{1}{l}{\textbf{67.51 $\pm$ 0.10}}\tabularnewline
Ours  & \textbf{62.63 $\pm$ 0.50}  & \multicolumn{1}{l}{\textbf{52.22 $\pm$ 0.12}} & \multicolumn{1}{l}{\textbf{73.76 $\pm$ 0.41}} & 67.36 $\pm$ 0.10 \tabularnewline
\bottomrule
\end{tabular}} \label{tab:table10way} 
\end{table*}

\begin{table*}[t]
\centering \caption{Evaluation of 15-way few-shot classification on MiniImageNet.}
\resizebox{.7\linewidth}{!}{%
\begin{tabular}{lcccc}
\toprule 
    \multicolumn{1}{c}{} & \multicolumn{4}{c}{MiniImageNet (15-way)}\tabularnewline
\midrule 
Model & 1-Shot Val  & 1-Shot Test  & 5-Shot Val  & 5-Shot Test \tabularnewline
\midrule 
PN \cite{snell2017}  & 45.64 $\pm$ 0.43  & 37.42 $\pm$ 0.10  & 63.47 $\pm$ 0.28  & 58.40 $\pm$ 0.08 \tabularnewline
AM3 \cite{xing2019}  & 49.23 $\pm$ 0.44  & 39.10 $\pm$ 0.10  & 66.06 $\pm$ 0.30  & \textbf{60.27 $\pm$ 0.08} \tabularnewline
Ours  & \multicolumn{1}{l}{\textbf{55.24 $\pm$ 0.35}} & \multicolumn{1}{l}{\textbf{43.54 $\pm$ 0.09}} & \multicolumn{1}{l}{\textbf{66.91 $\pm$ 0.28}} & 59.72 $\pm$0.08 \tabularnewline
\bottomrule
\end{tabular}} \label{tab:table15way} 
\end{table*}

\mypar{Model complexity} We report a comparison between AM3~\cite{xing2019} and our
approach in terms of the number of parameters, and inference latency in 
Table~\ref{tab:tableTime}. 
The number of parameters, shown in the second column, compares
the original AM3 model versus our complete model, excluding the backbone. 
Despite the addition of the new attention mechanisms, the final model has fewer parameters compared to the original AM3 model due to the simplification of $\alpha$ coefficients in our model, as explained above.
The latency values, shown in the following columns, are measured in terms of the average time it takes to complete a
single forward pass of a 5-way episode on a single A100 GPU, in the $1$-shot and $5$-shot settings.
The increased inference time is a result of the attention mechanisms.

\begin{table*}[t]
\centering
    \caption{The number of parameters (excluding backbone)
and the average forward pass duration for 5-way prediction on mini-ImageNet.}
\resizebox{.65\linewidth}{!}{
\begin{tabular}{cccc}
\toprule
Model & \multicolumn{1}{l}{Num. parameters} & \multicolumn{1}{l}{$1$-shot latency} & \multicolumn{1}{l}{$5$-shot latency} \\ \midrule
    AM3~\cite{xing2019}   & 475.541                              & 3 ms                                  & 4 ms                                  \\
Ours  & 345.945                              & 17 ms                                & 19 ms                               \\ \bottomrule
\end{tabular}}
\label{tab:tableTime}
\end{table*}

\begin{table*}[t]
\centering
\caption{Evaluation of sample attention for handling noisy support samples.}
\resizebox{.45\linewidth}{!}{%
\begin{tabular}{lc}
\toprule
Method                & \multicolumn{1}{l}{5-way 5-Shot Acc.}                 \\ \hline
PN  \cite{snell2017}            & 62.61 +- 0.22                         \\
Only Sample Attention                 & 70.48 +- 0.20                         \\ 
\bottomrule
\end{tabular}
}
\label{tab:tablenoisy}
\end{table*}

\subsection{Few-shot learning with noisy samples} 

While the problem of having support sample noise is not a commonly studied problem in FSL, possibly
due to the presumption that few samples are likely to be correct, there can be various real-world scenarios where
few-shot learner needs to operate over noisy support examples, \eg learning in
robotic systems through visual demonstration or through on-the-fly crawled samples of a desired class.
In this sense, FSL techniques can be interpreted in a broader context as fundamental
learning mechanisms as parts of approaches tackling noisy supervision problems.
In addition, the support sample noise setting also provides a challenging experimental setup for
understanding the effectiveness of the proposed sample attention mechanism.

Our sample noise experiments focus on the 5-shot 5-way setting on MiniImageNet.  We
artificially introduce noise into the support batches during training and testing using the
following procedure: for each class in a task, we guarantee the existence of at least three
correctly labeled support examples.  With 50\% probability, we apply label noise to the remaining
samples by randomly permuting their class labels, resulting in up to two noisy samples per class,
per task.  We note that such within-task label noise is more challenging and practically more
relevant than having completely irrelevant support samples, as unrelated samples are less likely to
cause cross-class confusion in the few-shot classification results. To evaluate sample attention in
an isolated manner, we use sample attention directly on top of PN in these experiments.

\begin{figure*}[t]
    \begin{center}
  \includegraphics[width=.65\linewidth]{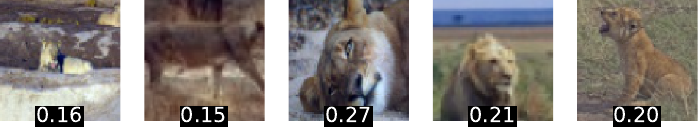} \\
  \includegraphics[width=.65\linewidth]{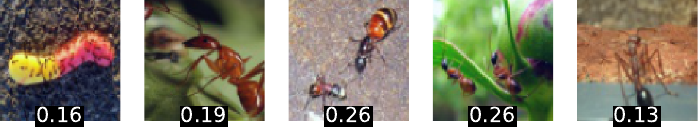} \\
  \includegraphics[trim=.68cm .5cm 0 0, clip, width=.65\linewidth]{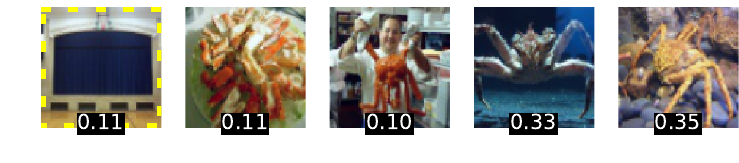} \\
  \includegraphics[trim=.68cm .5cm 0 0, clip, width=.65\linewidth]{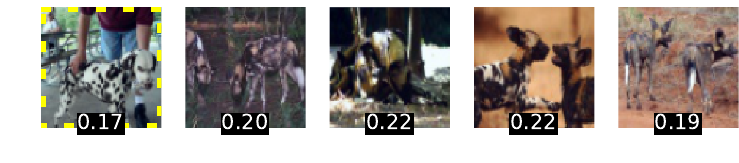}
    \end{center}
    \caption{Sample attention examples for clean (top two rows) and noisy (lower two rows) support sample settings. Yellow dashed lines represent the noisy samples. Numbers indicate attention scores.\label{fig:sampleattweights}}
\end{figure*}

\begin{figure}[t]
    \centering
    \includegraphics[width=0.65\linewidth]{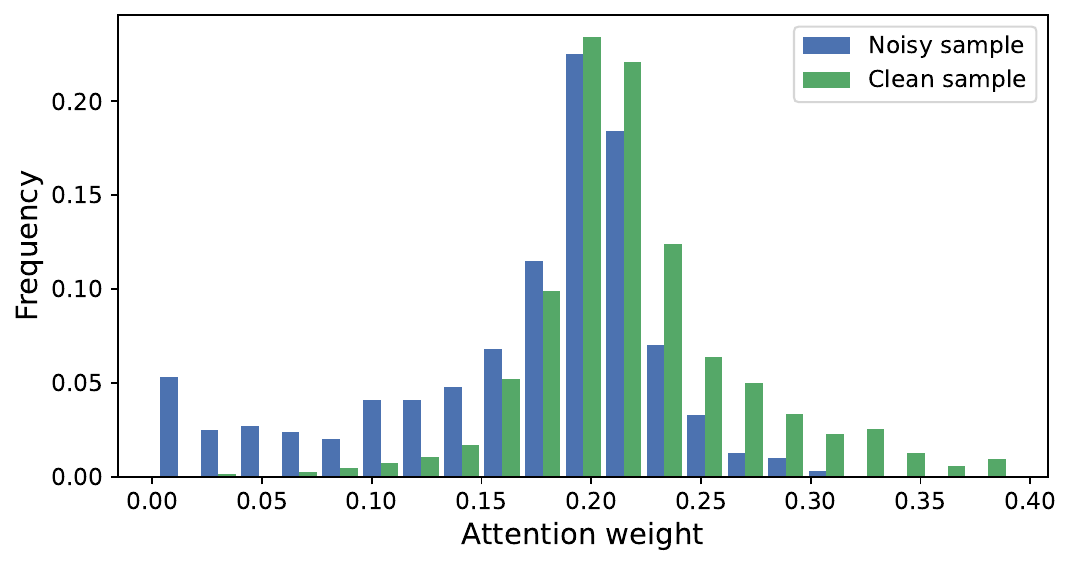}
    \caption{Distribution of sample attention weights for noisy and clean support samples.}
    \label{fig:fig_sampleatt_noisyhist}
\end{figure}

We report the results in Table~\ref{tab:tablenoisy}. We observe a large performance margin of
approximately $7.8$ points between the baseline ($62.61$) and noise-aware sample attention
($70.48$). This strongly suggests the ability of sample attention to utilize semantic priors for
selecting the most informative support samples. To better understand how sample attention improves
the results here, we also provide qualitative results both with and without noise in support
samples, in \fig{sampleattweights}. Each image 5-tuple corresponds to a 5-shot support set of a
class.  For comparison, we present the sample attention results from the clean support settings
(the upper two rows) and the noisy support settings (the lower two rows).  The dashed borders
indicate noisy samples, and the numbers indicate the resulting attention values. In the top-most
example, we observe that the model puts higher attention scores to support examples where the target
{\em lion} is most recognizable and estimates significantly lower attention to the images where the
lion is small (the first example in the row) or pretty much unrecognizable (the second one).
Similarly, in the second row, we observe that the model attends more strongly to the support samples
where the target {\em ant} instances are clearly recognizable. The example on the third row shows
that the model estimates a low attention score for the noisy (the first one in the row) and cluttered
(the second and third examples for the class) inputs for the target {\em king crab} classes. The
fourth row shows an example where the attention to the noisy sample is comparable to
the correct ones due to the similarity across the noisy {\em dalmatian dog} sample and the
target {\em African wild dog} class.

\mypar{Sample attention distributions} The quantitative and qualitative examples highlight that
sample, the proposed attention mechanisms can leverage semantics to a large extent in
prioritizing the training samples, but it can also unsurprisingly make (relative) mistakes due to
variety of reasons. To understand the {\em correctness} of the attention estimates in a quantitative
manner, beyond the accuracy values, we present the distribution of the attention score estimates for
clean and noisy samples in \fig{fig_sampleatt_noisyhist}.  Each histogram bin corresponds to
an attention score range and the shown frequencies are obtained over $100$ randomly sampled 5-way,
5-shot tasks with noisy support batches. In these two distributions, we make the following
observations: (i) attention scores given to noisy samples tend to be lower, especially where lower
than $0.10$ attention values are assigned almost exclusively to noisy samples, (ii) noise {\em
detection} has room for improvement, as the assignment of relatively higher attention scores to
noise samples are not rare, and (iii) the learned importance estimator works in a non-degenerative way
where attention scores vary from almost-zero values to $0.35$ and above values.

\section{Conclusions}
\label{sec:conclusions}

This paper proposes an approach for leveraging semantic prior knowledge in few-shot learning
of classifiers. The primary motivation of our approach is to utilize semantic prior knowledge about
classes to estimate the importance of support samples and representation dimensions. 
Our method performs well against the approaches that use only visual data and the ones
that use additional semantic information. The performance gains are more prominent in lower shot
settings where the need for auxiliary semantic information is typically higher.  We also study the
case of sample noise in support sets. We also investigate the applicability of our approach in a noisy FSL setting, where some
of the few-shot support samples are stochastically corrupted.  While we focus on a single semantic
modality throughout our experiments, we believe that incorporating multiple modalities, as in
\cite{schwartz2019}, integration into a generative FSL model-based approach, as in \cite{fu2019} 
and integrating self-supervised training tasks as in \cite{iept} are promising future work directions.

\section*{Acknowledgements}

This work was supported in part by the TUBITAK Grants 116E445 and 119E597.  The numerical
calculations reported in this paper were partially performed at TUBITAK ULAKBIM, High Performance
and Grid Computing Center (TRUBA resources).

\bibliography{abbr,mybib}

\end{document}